\documentclass[letterpaper, 10 pt, conference]{ieeeconf}  
\IEEEoverridecommandlockouts                              
\usepackage[dvipsnames]{xcolor}
\usepackage{graphicx}
\usepackage{cite}
\usepackage{graphics} 
\usepackage{epsfig} 
\usepackage{times} 
\usepackage{amsmath} 
\usepackage{amssymb}  
\usepackage{subcaption}
\usepackage[ruled]{algorithm2e}
\usepackage{siunitx} 
\usepackage{booktabs} 
\usepackage{makecell}
\usepackage{multirow}
\usepackage{color, colortbl}
\usepackage{adjustbox}
\usepackage{esvect}
\usepackage{tikz}

\definecolor{Gray}{gray}{0.9}

\newcommand{\mb}{\mathbf}
\newcommand{\etal}{\textit{et al}. }
\graphicspath{{fig/}}
\title{\LARGE \bf ROW-SLAM: Under-Canopy Cornfield Semantic SLAM}
\author{Jiacheng Yuan$^{1}$, Jungseok Hong$^{2}$, Junaed Sattar$^{2}$, and Volkan Isler$^{2}$
\thanks{$^{1}$ is with Department of Electrical and Computer Engineering,
        University of Minnesota, Minneapolis, MN, 55455, USA
        {\tt\small yuanx320@umn.edu}}%
\thanks{$^{2}$ are with the Department of Computer Science and Engineering, University of Minnesota, Minneapolis, MN, 55455, USA
        {\tt\small \{jungseok, junaed, isler\}@umn.edu}}%
}

\begin{document}


\maketitle
\thispagestyle{empty}
\pagestyle{empty}

\begin{abstract}
We study a semantic SLAM problem faced by a robot tasked with autonomous weeding under the corn canopy. The goal is to detect corn stalks and localize them in a global coordinate frame. This is a challenging setup for existing algorithms because there is very little space between the camera and the plants, and the camera motion is primarily restricted to be along the row. To overcome these challenges, we present a multi-camera system where a side camera (facing the plants) is used for detection whereas front and back cameras are used for motion estimation. Next, we show how semantic features in the environment (corn stalks, ground, and crop planes) can be used to develop a robust semantic SLAM solution and present results from field trials performed throughout the growing season across various cornfields. 
\end{abstract}
	
\section{Introduction}
Cornfield weed control conventionally has relied heavily on herbicides which are undesirable due to environmental and health-related concerns and can not be used in organic fields. 
The alternative, manual weeding, is labor-intensive and costly. Therefore, there has been significant interest in robotic weeding
~\cite{robotic-weeding-1, robotic-weeding-2, robotic-weeding-lettuce, robotic-weeding-laser}. 
However, most existing techniques focus on early season weeding. During this time, since the canopy is not closed, usually a reliable GPS signal is available. Further, the robots can obtain top-down views, which are convenient for detection and localization. 
In this work, we focus on mid-season weeding where the robot must operate under the canopy.

\begin{figure}[ht!]
        \centering
        \includegraphics[width=0.8\linewidth]{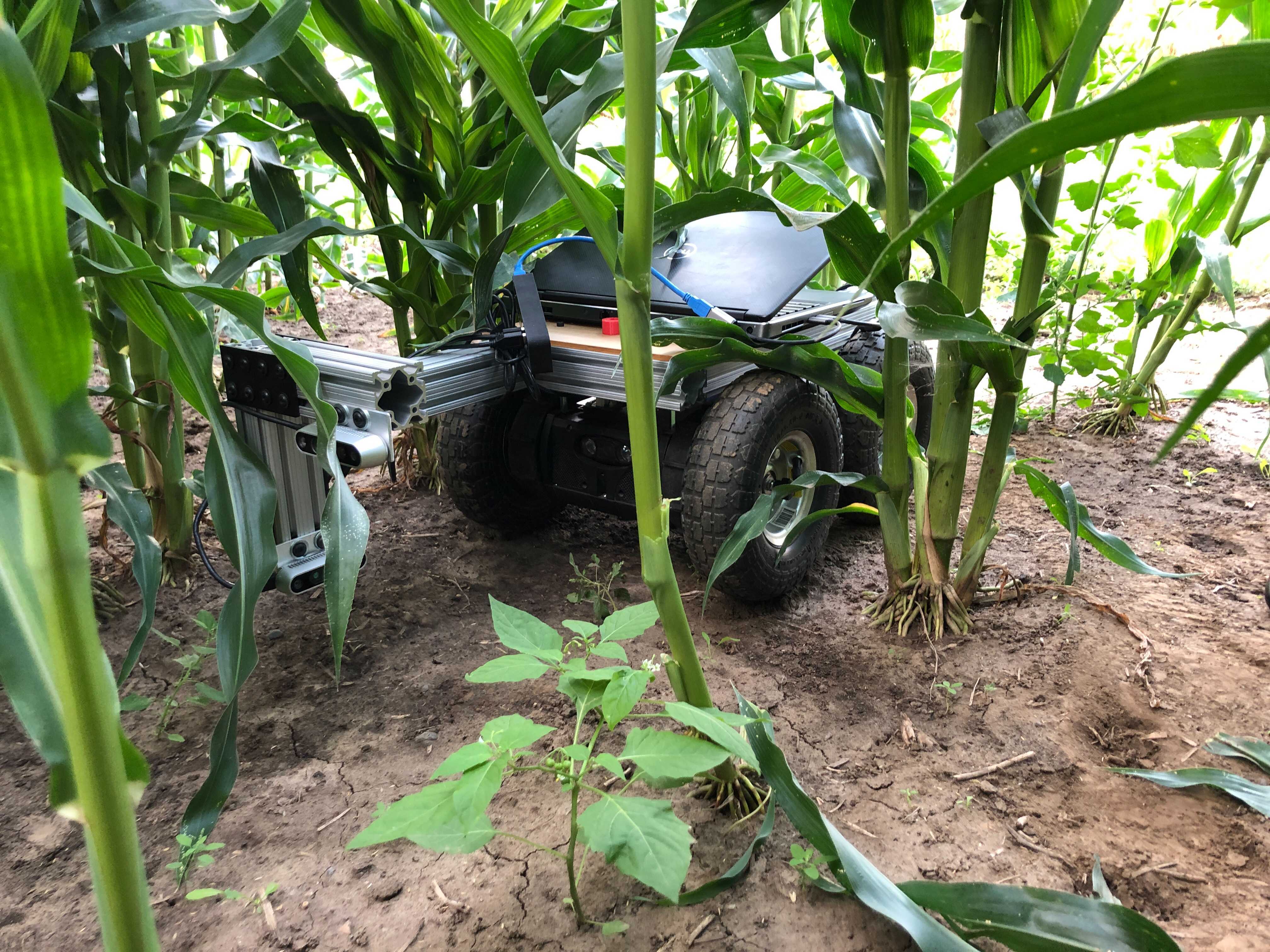}
        \caption{\small{Our robot in a mid-season corn field. Dense canopy and narrow row spacing makes it challenging for semantic SLAM.}}
        \label{fig:intro}
            \vspace{-6mm}
    \end{figure}

This setup introduces unique challenges for autonomous weeding: (1) top-down views are no longer available, and therefore a ``planar world" assumption does not hold (2) mid-season corn and weed canopies are usually in close distance and overlap with each other (3) there is frequent occlusion from corn leaves and weed even if we lower the height of the cameras closer to the ground in between two rows. (Fig.~\ref{fig:intro}) 

We observe that established SLAM algorithms~\cite{detect-slam, object-slam-2} struggle in the setup because: 
(1) dynamic features such as shaking leaves and weeds due to wind or robot motion cause the SLAM module to fail (2) the corn row lacks regular structures such as cylinders or walls, making it challenging to observe the corn plane from the map directly.

\begin{figure}
	    \vspace{0.2cm}
     \centering
     \begin{subfigure}[b]{0.235\textwidth}
         \centering
         \includegraphics[width=\textwidth]{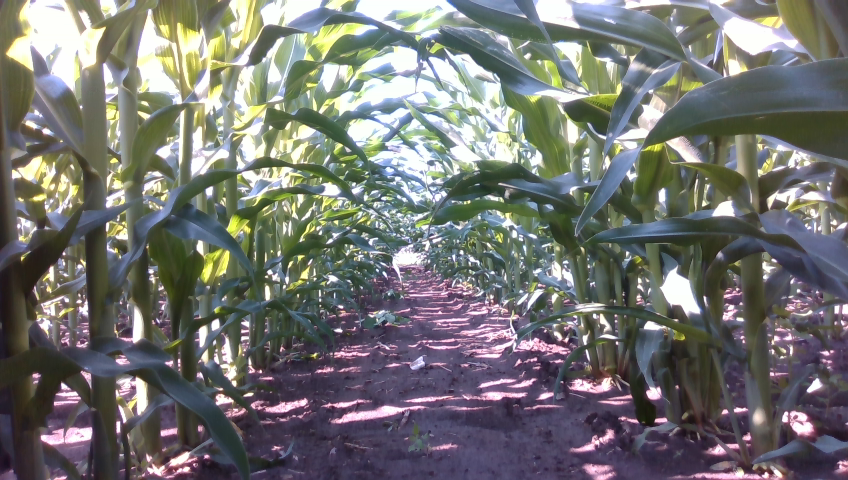}
                 \vspace{-5mm}
         \caption{\small{Front view (conventional)}}
     \end{subfigure}
     \hfill
     \begin{subfigure}[b]{0.235\textwidth}
         \centering
         \includegraphics[width=\textwidth]{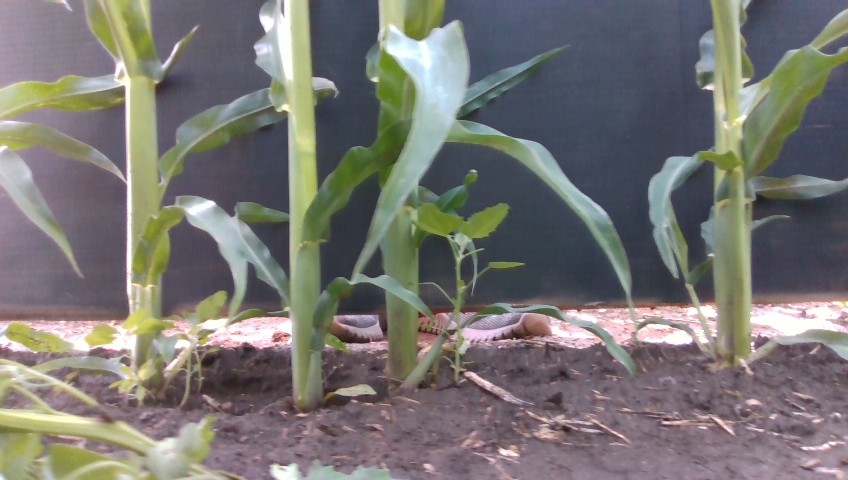}
                          \vspace{-5mm}
         \caption{\small{Side view (conventional)}}
     \end{subfigure}
     
     \begin{subfigure}[b]{0.235\textwidth}
         \centering
         \includegraphics[width=\textwidth]{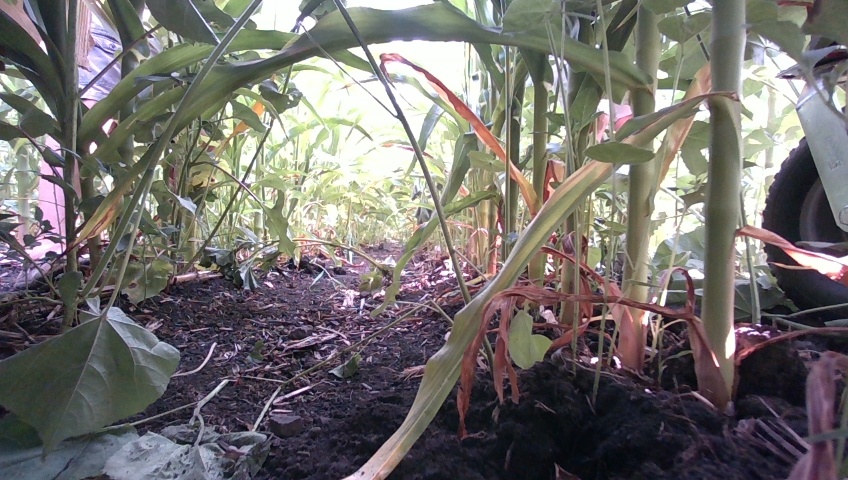}
                          \vspace{-5mm}
         \caption{\small{Front view (organic)}}
     \end{subfigure}
     \hfill
     \begin{subfigure}[b]{0.235\textwidth}
         \centering
         \includegraphics[width=\textwidth]{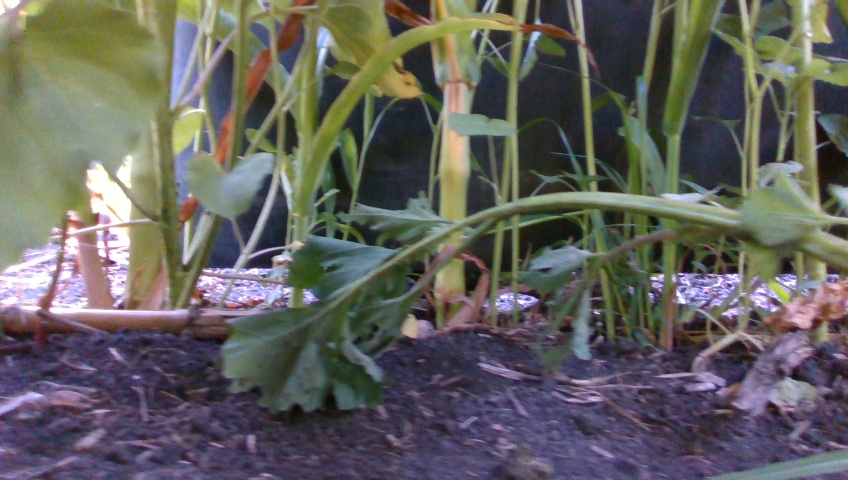}
                 \vspace{-5mm}
         \caption{\small{Side view (organic)}}
     \end{subfigure}
     
    \caption{\small{Sampled images from both the conventional field and the organic field. The conventional field has almost no weeds due to herbicides, while the organic field shows heavy weed infestation.}}
    \label{fig:sample_images}
        \vspace{-6mm}
\end{figure}

In Fig.~\ref{fig:sample_images}, we present image samples from conventional and organic fields with heavy weed infestation as a comparison. We highlight the stronger interference from the weed stems and lower corn stalk visibility due to the existence of the weeds.

In this paper, we introduce a multi-view vision system for the detection and 3D localization of corn stalks in mid-season corn rows, called ROW SLAM. We present this system (Fig.~\ref{fig:intro}, \ref{fig:diagram}) as a prototype of the vision module for an autonomous weeding robot currently under development. In our approach, we model the corn stalks in the same row as a plane that is perpendicular to the ground. A ground-view camera (back camera) performs SLAM using ground plane features to obtain 3D odometry. We implement a Structure-from-Motion (SfM) strategy that accommodates the multi-view inputs to estimate the 3D pose of the corn stalk plane. Combining this motion estimation module with the object detection and tracking modules, we build a map that has both metric (location and orientation) and semantic information of the corn stalks in it. We present field results which demonstrate the accuracy and robustness of our approach toward weeding mid-season cornfields.

Our contributions can be summarized as follows:
\begin{itemize}
    \item We present \textbf{ROW SLAM}: a multi-camera vision system with non-overlapping views for semantic SLAM in mid-season corn rows. Specifically,  we use the system to build maps with both 3D location and semantic information of corn stalks.
    \item We present a multi-view Structure-from-Motion (SfM) strategy for robust corn stalk plane estimation.
    \item We test our system in 8 different corn rows with heavy weed infestation across the growing season. Our method outperforms all baseline approaches.
\end{itemize}

\section{Related Work}

	\begin{figure*}
    
	    \vspace{0.2cm}
    \centering
    \scalebox{0.75}{\input{fig/diagram}}
    \caption{(\textbf{Top}) Proposed pipeline of ROW SLAM: Our pipeline takes images from \{front, back, side\} cameras and yields 1) IDs for each corn (denoted with $C_{\{a,..,e\}}$ in the scene, and 2) 3D positions of the corn stalks (denoted by yellow cylinders). (\textbf{Bottom-Left}) Proposed robot design with multi-view cameras. (\textbf{Bottom-Right}) A 3D reconstruction sample of the corn field. Beige and green planes represent ground and corn planes, respectively. Each camera pose is displayed separately.}  
    \label{fig:diagram}
    \vspace{-4mm}
\end{figure*}
The topic of robotic weed control~\cite{pandey2020aliterature} has been extensively studied over the last decade from various viewpoints: weed detection, field coverage, and vehicle development. Most recent weed control robots have been developed in the form of drones and ground vehicles. While drone-based approaches~\cite{uavweed} can cover large areas, they only provide weed information to manage weeds via ground vehicles (\textit{e.g.}, applying herbicide)~\cite{castaldi2017assessing}. Recently, Wu \etal~\cite{wu2020robotic} proposed a method using ground vehicles with downward-looking cameras to recognize weeds and remove them by end effectors. However, such an approach could fail as the crop canopies grow and block the top view.

Sivakumar \etal~\cite{under-canopy-navigation} use a front view camera to predict a vanishing point and corn lines for navigation under-canopy tasks. However, using the front view loses a lot of the details of the objects in the narrow corn row due to the viewing angle. In Sec.~\ref{sec:baselines}, we compare against their method for corn plane estimation. Zhang \etal~\cite{corn-stand-counting} use side view for corn stalk counting, but we show that side view also experiences frequent occlusion. To get accurate semantic features (\textit{i.e.}, corn stalks, ground plane, corn plane) from the map, we propose to combine the input from multiple views.


SfM~\cite{moulon2013global} is one of the classical methods available to obtain the 3D location of the features. In our application, since there is very little space and frequent occlusion between the camera and the plants, the camera motion is also restricted. Classical SfM strategies perform poorly for estimating the motion when observing only the plants. Hasheminasab \etal~\cite{gps-sfm} proposes to assist the SfM pipeline with GPS signals. Since GPS is unreliable under the canopy, we use SLAM odometry from the back camera and the front view corridor estimation to assist the SfM process using the side view.

For localization in outdoor environments, SLAM with RGB input~\cite{rgbd-slam, orb-slam, rtab-map} has been widely used. However, the localization accuracy of SLAM is sensitive to dynamic features. 
Recent works~\cite{detect-slam, slam-dynamic-scene} adopt neural networks to detect moving objects and remove their features. The remaining static features are then used to build the map.  
In our application scenario, most of the features above ground are susceptible to unconstrained motion due to the interference from wind or robot motion, making it challenging to detect dynamic features. So we select the ground view as input for the SLAM module to ensure the static map assumption. In Sec.~\ref{sec:results} we compare the performance using different views to support this choice.

Another aspect of obtaining semantic features is object detection.
Some recent works~\cite{approximation_model_1, approximation_model_2, object-slam-2} build offline approximation models for the objects of interest and detect them by model matching while building the map. However, such a method is unsuitable for corn stalk detection due to the lack of structured shapes in the non-uniform canopy. The dense occlusion from the leaves and the weeds makes it even more challenging to detect corn stalks by matching offline models.

Recent advances in deep learning allow accurate real-time object detection~\cite{zou2019object} and detect a wide range of objects from a single model with a large amount of data.
Such advances have enabled adopting deep learning-based object detection in agricultural applications~\cite{KAMILARIS201870}, and many studies~\cite{hasan2021survey} apply the object detection to detect weed. However, these approaches are limited because they (1) require a large amount of weed data to train the models which is challenging to obtain, (2) can only be used for the field that existing weed information is previously known, and
(3) may need a survey of a target field to collect field-specific weed information and its visual data before the deployment in a new field. 
In contrast, we propose to focus on detecting corns and consider the remaining as weeds. Our approach can be applied to a broader range of fields and growing stages since corns have fewer variants compared to weeds.

To associate object detection across frames, object tracking algorithms~\cite{ciaparrone2020deep} with visual sensors have been proposed to track a wide range of objects (\textit{e.g.}, pedestrians~\cite{brunetti2018computer}, vehicles~\cite{osep2017combined}, sports players~\cite{bridgeman2019multi}, crops~\cite{liu2018robust, appletrack}). The tracking algorithms generally use an estimation model and data association method such as optical flow~\cite{horn1981determining}, Kalman filter, and Hungarian algorithm to associate detection results from the previous frame to the next frame. In our pipeline, we use Simple online and realtime tracking (SORT)~\cite{bewley2016simple} due to its speed and accuracy, as demonstrated on the MOT15~\cite{leal2015motchallenge} dataset.

In this work, we show that robotic weed control in mid-season corn row poses unique challenges to existing SLAM algorithms. With our system design and combining multiple existing algorithms, the results demonstrate that we can overcome the challenges.


\section{Problem Formulation}
We are given a cornfield planted in a row, and have a vehicle that can traverse the field. A camera frame is rigidly attached to the vehicle and moves under the corn canopy with three cameras (one in the front-facing along the row, one on the side facing the corn plane, and one on the back facing the ground as shown in Fig.~\ref{fig:diagram}). 

Our goal is to detect the corn stalks and localize them in the global coordinate frame. Therefore, we formulate this problem as a semantic SLAM problem where we aim to build a map with semantic features (corn stalks, corn plane, and the ground plane). We choose corn stalks as the target semantic feature instead of weeds because, unlike the corn stalks, weeds in the field have various species and appearances. They also tend to vary across fields and regions. As a result, corn detection can be more consistent and robust than weed detection. Once corn stalks are identified, we can treat the remaining plants as weeds.

\section{System Overview}
In this section, we introduce the necessary mathematical notations in Table~\ref{tab:notations} and describe our system's hardware. We discuss the details of our approach for 3D corn stalk detection and localization next.

	\subsection{System Description}
	

	
    Our hardware system consists of three cameras (Fig.~\ref{fig:diagram} Bottom-Left). 
    The front and side cameras (Intel Realsense D435) publish RGB images and depth images at $30$ Hz. The back camera (Intel Realsense D435i) has a built-in IMU and is mounted at an angle facing the ground. The back camera publishes RGB and depth images at $30$ Hz, gyro at $400$ Hz, and accelerometer at $250$ Hz. We use ROS for robot control and routing sensor data. 
    For our experiments, we mount the system on a small 4-wheel rover from Rover Robotics.

    \begin{table}[t!]
    \centering
    \caption{\small{Summary of Notations}}
    \label{tab:notations}
    \begin{tabular}{l l} 
        Notation & Description  \\
        \midrule
        $\mb n_p, d_{p}$   & \makecell*[{{p{6.5cm}}}]{Normal vector and distance to the origin for the corn plane (shown in Fig.~\ref{fig:diagram}) } \\
        $\mb n_g$ & \makecell*[{{p{6.5cm}}}]{ Normal vector for the ground plane  (shown in Fig.~\ref{fig:diagram})}  \\ 
        $\mb v_l$          & \makecell[{{p{6.5cm}}}]{Intersection of the ground plane and the corn plane, also the orientation of the corn row} \\
        \midrule
        $\mb T$ & \makecell*[{{p{6.5cm}}}]{Relative transformation between sequential side camera poses}\\
        \bottomrule

    \end{tabular}
\emph{Notes:} bold uppercase letters for matrix, bold lowercase for column vector, normal ones are scalar
              \vspace{-6mm}
\end{table}

	\subsection{Method Overview}
    As shown in Fig.~\ref{fig:diagram}, we use RGB images from the side view camera for corn stalk detection. The detections only localize the stalks in 2D. Thus, to find the 3D position of the detected corn stalks, we also need to find the 3D pose of the corn plane. We address the corn plane estimation problem in two parts: (1) finding the plane normal direction, $\mb n_{p}$, and (2) finding the distance from the camera center to the plane, $d_{p}$, shown in Fig.~\ref{fig:system}.
    
    We use the front and back camera to assist with the estimation of the corn plane $\mb n_{p}$ and $d_p$. With the RGB-D input from the front view camera, we estimate the ground plane normal direction $\mb n_g$ and the direction of the corn line $\mb v_{l}$. The corn plane orientation is found by the cross product of these two vectors.
	\begin{equation}
	    \label{corn_line}
	  \mb n_{p} = \mb n_{g} \times \mb v_{l}  
	\end{equation}
	
	Then, we use the corn plane normal $\mb n_p$, combined with side view object detection results and SLAM odometry $\mb T$ as the input for the Multiview SfM module to estimate $d_p$.
	Before going through the details of this module, we first introduce the detection and tracking.
	
	\subsection{Corn Stalk Detection}
    We implement our corn stalk detection model using Faster R-CNN~\cite{ren2015faster} with MobileNet V3~\cite{howard2019searching} as the backbone and Fully Connected Network (FCN) as a head. We select MobileNet to provide fast inference, thus preventing the deep detection model from being a bottleneck in our pipeline. The backbone can be replaced with a larger network such as Resnet-18, -34, -50~\cite{he2016deep} to improve the accuracy of the model.
    We train our model with pre-trained weights using the COCO dataset~\cite{lin2014microsoft}, and refine further using our corn stalk dataset. 
	\subsection{Corn Stalk Tracking}
	With the outputs from the detection model, we implement the corn stalk tracking pipeline using (1) optical flow with centroids~\cite{revathi2012certain}, and (2) simple online and realtime tracking (SORT)~\cite{bewley2016simple}. 
	\subsubsection{Optical Flow}
	We build the corn stalk tracking algorithm by applying optical flow algorithm and the corn detection results. The model detects bounding boxes for each corn stalk every 200 frames, and the centroids of the bounding boxes are calculated. After we obtain the centroids from the output of the detection model at frame $X$, the iterative Lucas-Kanade method with pyramids~\cite{bouguet2001pyramidal} is used to track each centroid using optical flow from frame $\{X+1\}$ to $\{X+199\}$.
	
	\subsubsection{SORT}
	SORT performs four actions for each input frame: detection, estimation, data association, and update tracking identities. 
	The detections from Faster R-CNN are propagated to the next frame using a linear constant velocity model. The results are also utilized to update the target state with the Kalman filter~\cite{welch1995introduction}. When a new detection is obtained from the next frame, the Hungarian algorithm~\cite{Kuhn55thehungarian} is used to associate detections from the previous frame to the current frame using the intersection-over-union (IoU) metric. When the IoU metric is below a predefined threshold, then new identification of the detection is created. 

	\begin{figure}[t!]
	    \vspace{0.2cm}
        \centering
        \includegraphics[width=0.6\linewidth]{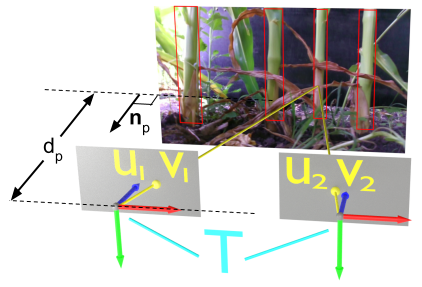}
        \caption{\small{Illustration for the multi-view SfM strategy. The relative camera pose (mint) is given by the SLAM module, and feature matching is masked by detected bounding boxes. $\mb n_p$ is the plane normal of the corn plane, and $d_p$ is the distance between the corn plane and the left camera center.}}
        \label{fig:system}
            \vspace{-4mm}
    \end{figure}

         
     
	\subsection{Corn Plane Estimation}
	\textbf{Plane Normal} 
	To get the ground plane pose, we first use the front view RGB-D input to compute the point cloud of the scene, followed by color thresholding and RANSAC~\cite{ransac} plane fitting. The points are usually denser when closer to the camera center, so we downsample the ground plane inlier points before applying principal component analysis (PCA). Since the corn row is narrow ($\approx 70cm$), the two eigenvectors with the largest and smallest eigenvalues from PCA indicate the directions of the corn line $\mb v_{l}$ and the ground plane normal $\mb n_{g}$, respectively. The corn plane normal is then computed by the cross product of these two vectors.
	
	\textbf{Plane Distance} Across different frames, the motion of the features on the corn stalks can be modeled by planar homography. However, instead of doing the full SfM purely on the side view input, we find it much more stable if we use the ground features to estimate the camera trajectory. We thus use the images from the back camera, providing a predominantly ground-plane view to avoid the unstable features from the corn leaves and weeds. With these images, we use the off-the-shelf RTAB-Map SLAM \cite{rtab-map} package to perform visual SLAM and use the Sigma Point Kalman Filter \cite{spkf} to fuse the SLAM odometry with IMU input. The resulting odometry runs at $200$ Hz. After calibrating the back view and side view cameras, we can use the camera trajectory as is. Combining this with the estimated corn plane normal, we formulate plane distance estimation as the following least square problem over re-projection error (shown in Fig.~\ref{fig:diagram} as the Multiview SfM block):
	
	Given two side view RGB frames $I_1, I_2$, the relative transformation $\mb T_1^2$ between their camera poses (where $ \mb R$ and $\mb c$ is the rotation and translation component) and the corn plane normal $\mb n_{p}$, we want to estimate the plane distance $d_{p}$, as shown in Fig.~\ref{fig:system}.
	
	We first run Faster R-CNN on $I_1, I_2$ to get bounding boxes for detected corn stalks. Next, we compute SIFT features $f_1, f_2$ for $I_1, I_2$  only within the regions defined by the bounding boxes, and then use cross-matching and ratio test to find good matches between  $f_1, f_2$. For each matched pair, let $(u_1, v_1)$ and $(u_2, v_2)$ be the corresponding pixel coordinates, $\mb K$ and $\lambda_1$ respectively be the camera intrinsic matrix and the depth for the first feature. 
	
	Letting $\mb x_1 = \begin{bmatrix} u_1 & v_1 & 1 \end{bmatrix}^T$, the 3D position of $f_1$ can be expressed as: $\lambda_1   \mb K^{-1} \mb x_1$. 
			 
	Since we assume the feature lies in the corn plane, according to the plane equation 
	\begin{equation}
	    \lambda_1 \mb n_{p}^T  \mb K^{-1} \mb x_1 + d_{p} = 0
	\end{equation}
	
	We can rewrite $\lambda_1$ in terms of $d_p$:
	\begin{equation} \label{eq:depth}
	    \lambda_1 = - d_{p} / \mb n_{p}^T  \mb K^{-1} \mb x_1
	\end{equation}
	
	After applying the rotation $\mb R$ and translation $\mb c$, the projection of the 3D position of $f_1$ in the second camera frame can be expressed as:
	 \begin{equation} \label{eq:reprojected-3d}
	    \lambda_1 \mb K \mb R \mb K^{-1} \mb x_1 + \mb K \mb c
	 \end{equation}

	If we substitute Eq.~\ref{eq:depth} in Eq.~\ref{eq:reprojected-3d} we can rewrite Eq.~\ref{eq:reprojected-3d} as follows:
    \begin{equation}
        - d_{p} \begin{bmatrix} l_0 & l_1 & l_2 \end{bmatrix} + \begin{bmatrix} s_0 & s_1 & s_2 \end{bmatrix}
    \end{equation}
    where,
    \begin{align}
        \begin{bmatrix} l_0 & l_1 & l_2 \end{bmatrix}^T  &=  \frac{ - 1 }{\mb n_p^T \mb K^{-1} \mb x_1} \mb K \mb R \mb K^{-1} \mb x_1 \\
        \begin{bmatrix} s_0 & s_1 & s_2 \end{bmatrix}^T & =  \mb K \mb c
    \end{align}
    
    The projected pixel position of $f_1$ in the second camera $(u_2', v_2')$ is:
    \begin{align}
        \begin{cases}
            u_2' = (-d_{p} l_0 + s_0)/(-d_{p} l_2 + s_2) \\
            v_2' = (-d_{p} l_1 + s_1)/(-d_{p} l_2 + s_2)
        \end{cases}
    \end{align}
    
    We can linearize the re-projection error $(u_2-u_2', v_2-v_2')$ with respect to $d_{p}$ so it can be estimated by least squares and RANSAC. However, notice that the least square robustness is sensitive to the second term $\mb K \mb c$ in Eq.~\ref{eq:reprojected-3d}. So when applying this algorithm, we need to make sure the translation scale is above a threshold to ensure robust triangulation.
    
\subsection{3D Localization}

    The 3D position of a corn stalk is obtained by the projection of the bounding box centroid onto the corn stalk plane. To show that it is more robust than directly using RGB-D input, we compare against RANSAC Plane Fitting in Sec.~\ref{sec:results}. We also use the tracking module to link the 3D corn stalk positions with corn IDs. Such association across multiple frames allows us to reject outliers and false positives, making the localization more robust.
\section{Experiments and Results}
We present our data collection procedures, followed by the evaluation metrics for our method. Then we introduce the baseline methods we compare against.

	\begin{figure}[t!]
	    \vspace{0.2cm}
        \centering
        \includegraphics[width=0.9\linewidth]{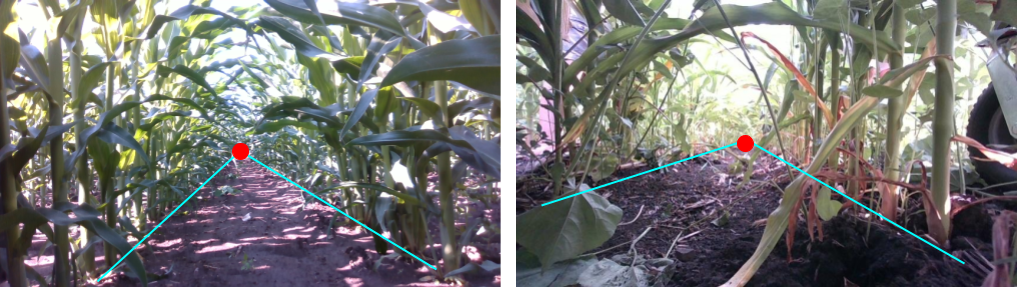}
        \caption{\small{Corridor projection of the corn row. Two corn lines (blue) are parallel and intersects at the vanishing point (red). Due to occlusion, the corn lines are usually not clearly visible.}}
        \label{fig:vanishing-line}
                    \vspace{-4mm}
    \end{figure}
\subsection{Data Collection}

The rover is controlled by a human operator to drive at around $0.3m/s$. 
To provide additional control of imaging conditions, we used a wheeled arch platform to go over the rows and provide cover for the rover. In the future, we are planning to combine the rover and the cover into a single robotic platform.  Our collected dataset includes $8$ different corn rows where $4$ of them are from conventional fields, and the other $4$ are from organic fields. We collected data across different growing stages throughout the mid-season (between V$5$ stage and V$12$ stage, late June to early August in Minnesota, USA). 


\begin{figure*}[t!]
	    \vspace{0.2cm}
    \centering
    \includegraphics[width=0.85\linewidth]{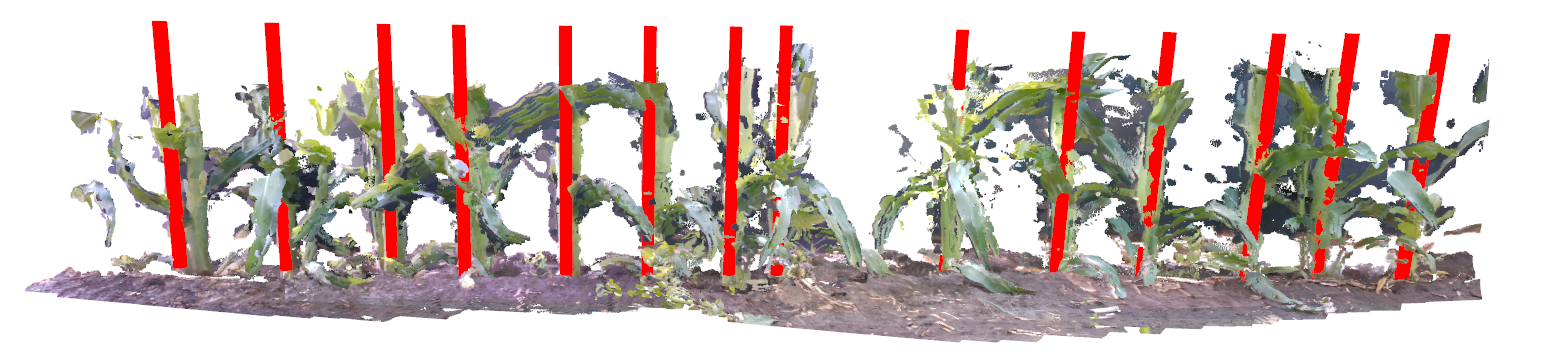}
    \caption{\small{The reconstructed point cloud of a corn row and the estimated corn stalk position (red cylinders) from ROW SLAM. }}
    \label{fig:side_recon}
                \vspace{-4mm}
\end{figure*}
	
\subsection{Evaluation Metric}
We perform offline evaluations using the dataset collected in two different aspects, detection accuracy and localization. For evaluating the corn detection module, we label $763$ images and separate them into $563$ training images ($235$ images from conventional field and $328$ images from organic field) and $200$ test images ($100$ images from each field). Training and test dataset images for each field are picked from different rows to reduce the correlation between the datasets and provide an accurate measure of the generalization capability of our network. For localization under the canopy, it is difficult to get high-accuracy GPS signals reliably. So we manually measure the distance between neighboring corn stalks and use the corn stalks as landmarks along a straight line to evaluate the localization accuracy. We first line fit on the camera trajectory and project predicted target positions to the fitted line to reduce their dimension to one. To remove the duplicated predictions for a single target, we compute its nearest neighbor from the world measurements for each predicted corn stalk position. The distance error $\epsilon_1$ is computed as the mean absolute error of the neighboring corn stalks distance between prediction and real-world measurements. Finally, we measure the tracking variance through re-projection error $\epsilon_2$. After accumulating the 3D position measurements for each corn stalk, we compute the mean position of the centroids and re-project them back to each frame. The re-projection error is computed by the mean pixel error between the centroids and the centroids of the bounding boxes proposed by the tracking module across all the frames.



\subsection{Baselines}
\label{sec:baselines}
\textbf{SLAM Input} 
As the SLAM pipeline is sensitive to moving objects in the scene, we compare the robustness and accuracy of SLAM while using different input views. Specifically, we replace the SLAM input in Fig.~\ref{fig:diagram} with the front camera view or side camera view and re-run the Semantic SLAM pipeline. 

\textbf{Corridor Prediction}
The corridor prediction method replaces the corn plane prediction module in Fig.~\ref{fig:system}. It is also based on the planar approximation of the corn row. Since the corn stalk planes on both sides are parallel, their intersection line with the ground plane must also be parallel. Therefore by projective geometry, they intersect in the image plane at the vanishing point (Fig.~\ref{fig:vanishing-line}). After obtaining the 3D pose of the ground plane, the vanishing point can be used to get $\mb v_{l}$. The corn stalk plane normal $\mb n_p$ can be computed by the cross product of $\mb v_l$ and $\mb n_g$. Then, $d_p$ can be obtained by the 3D position of any point along the corn line since it is also on the ground plane.

It is known that color thresholding yields unreliable results~\cite{under-canopy-navigation} for vanishing line detection. Thus, we combine the ResNet34 head with a 3-layer multi-layer perceptron to predict the vanishing point pixel location and two slopes for the corn line. We do not predict the full parameters for the two lines so that we can enforce the projective geometry.

\textbf{RANSAC Plane Fitting} 
The RANSAC plane fitting method replaces the multi-view SfM module in Fig.~\ref{fig:system} for the estimation of $d_p$.
After obtaining the side plane normal $\mb n_{p}$, one way to estimate $d_p$ is by direct observation of the 3D scene in the side view. Therefore to support our claim on the low observability of the corn plane, we present plane fitting in the side view as another baseline method. The detected object bounding boxes are used to find candidate points in the corn plane. We only need the 3D position of one selected point with the plane normal to compute $d_p$. Then RANSAC is applied to make the estimation robust.

\textbf{Optical Flow Tracking} 
To ensure high localization accuracy, we use SORT as our tracking module which runs detection at every frame and uses the Kalman Filter to make the tracking more robust. In this baseline, we replace the tracking module with optical flow tracking. Unlike SORT, the optical flow tracking method will re-initialize target id every $k$ frame. In our case, we choose $k=200$.

\begin{table}[t!]
	\centering
		\caption{\small{Faster RCNN Corn Detection Accuracy ($\%$)}}
		\label{tab:corn-detect}
		\begin{tabular}{l c c} 
			\toprule
			{} & \bf{Conventional field} & \bf{Organic field} \\
			\midrule
								   
			$AP$   &   26.2    & 47.8  \\
			$AP_{50}$ &  \bf{83.1}  &  \bf{89.1} \\
			$AP_{75}$ &  9.3 &  46.6\\
			\bottomrule
		\end{tabular}
    \label{tab:rcnn}
		            \vspace{-4mm}
\end{table}

\begin{table}[t!]
	\centering
	\captionsetup{justification=centering,margin=7mm}
		\caption{\small{Performance of Corn Stalk Localization ($\epsilon_1$: Metric Error, $\epsilon_2$: Centroid Re-Projection Error)}}
		\label{tab:corn-loc}
    		\begin{tabular}{l c c c c} 
    			\toprule
    			{} & \multicolumn{2}{c}{\bf{Conventional field}} & \multicolumn{2}{c}{\bf{Organic field}} \\
    			{} & $\epsilon_1$ (cm) & $\epsilon_2$ & $\epsilon_1$ (cm) & $\epsilon_2$ \\
    			\midrule
    								   
    			our approach           & \bf{1.8}    & \bf{5.6}  & \bf{3.5}   & \bf{10.4} \\
    			corridor prediction    &    8.5     &    19.3   & 9.7  & 27.8 \\
    			front-view SLAM        &     2.9     &     8.2   & 6.2  & 13.3  \\
    			side-view SLAM         &     4.5     &    16.8   & 7.1   & 15.5 \\
    			RANSAC plane fitting   &     3.7     &     9.5   & 9.3  & 24.2 \\
    			optical flow tracking  &     3.3     &    11.7   & 4.6  & 17.9 \\
    			\bottomrule
    		\end{tabular}
                \vspace{-4mm}
\end{table}

\subsection{Results}
\label{sec:results}
Faster R-CNN is trained for $100$ epochs, and Table~\ref{tab:corn-detect} shows evaluation results for each field. It tracks objects at $\approx 11$ FPS on a laptop (Intel i7-8850H, Quadro P3200, and 32GB RAM).
The organic field case has higher Average Precision (AP) values since about $60 \%$ of the training dataset is from organic fields. $AP_{50}$ ($AP$ at IoU=$.50$) has the highest accuracy for both cases, and it is because irregular shapes of the corns tend to have relatively low IoU values for many detections while they are still detected as corns. 

We compute the metric localization accuracy by comparing predicted target positions against the manually measured distance between corn stalks. We also evaluate the tracking variance through the re-projection error of the centroid. In Table~\ref{tab:corn-loc} we summarize the metric error $\epsilon_1$ and centroid re-projection error $\epsilon_2$ using the dataset from both conventional field and the more challenging organic field.

Our method outperforms all the baseline methods, which demonstrates its accuracy and robustness. As a comparison, the corridor prediction method shows fragile estimations of $d_p$. Due to the challenge of accurate estimation for $\vv{v}_l$ and projective geometry, the error of $d_p$ is significantly impacted by the angular error of the vanishing line in the front view. As for the view selection in SLAM, we notice that the front-view SLAM achieves comparable accuracy with our method during a windless day in the conventional field. However, in general, the comparison between our ROW SLAM and front/side view SLAM shows that the static features are critical for accurate under-canopy localization in the corn rows. For the RANSAC plane fitting method, the performance drop indicates the challenge in directly observing the corn stalk plane from a single view. Lastly, we observe greater centroid position drift in image plane with optical flow tracking than SORT, which affects the localization accuracy and variance, proving the importance of robust visual tracking. 

\section{Conclusions}
This paper presents our effort to address the under-canopy corn stalk detection and localization problem within narrow corn rows by proposing ROW SLAM and a multi-view camera system mounted on a ground vehicle. ROW SLAM combines existing algorithms (\textit{i.e.}, object tracking, SLAM, SfM) and applies several geometric methods to accurately estimate the three planes that bound the row as well as individual plants. Our results demonstrate ROW SLAM yields an accurate map containing corn stalk positions and their IDs while existing SLAM algorithms fail. Future work will add an end-effector to our existing system to remove weed physically. Furthermore, we are developing our algorithm to minimize the drift from the SLAM module by using corn stalks as landmarks together with an offline map.  


\bibliographystyle{IEEEtran}
\bibliography{weed.bib}

\end{document}